\documentclass[10pt,twocolumn,letterpaper]{article}

\usepackage{cvpr}
\usepackage{times}
\usepackage{epsfig}
\usepackage{graphicx}
\usepackage{amsmath}
\usepackage{amssymb}
\usepackage{dsfont}
\usepackage{comment}

\usepackage[pagebackref=true,breaklinks=true,letterpaper=true,colorlinks,bookmarks=false]{hyperref}

\cvprfinalcopy 

\def\cvprPaperID{4385} 

\ifcvprfinal\fi
\begin{document}

\title{Image-Question-Answer Synergistic Network for Visual Dialog}

\author{Dalu Guo, Chang Xu, Dacheng Tao\\
UBTECH Sydney AI Centre, School of Computer Science, FEIT, 
\\University of Sydney, Darlington, NSW 2008, Australia\\
\{\tt\small dguo8417@uni., c.xu@, dacheng.tao@\}sydney.edu.au
}

\maketitle

\begin{abstract}
The image, question (combined with the history for de-referencing), and the corresponding answer are three vital components of visual dialog. Classical visual dialog systems integrate the image, question, and history to search for or generate the best matched answer, and so, this approach significantly ignores the role of the answer. In this paper, we devise a novel image-question-answer synergistic network to value the role of the answer for precise visual dialog. We extend the traditional one-stage solution to a two-stage solution. In the first stage, candidate answers are coarsely scored according to their relevance to the image and question pair. Afterward, in the second stage, answers with high probability of being correct are re-ranked by synergizing with image and question. On the Visual Dialog v1.0 dataset, the proposed synergistic network boosts the discriminative visual dialog model to achieve a new state-of-the-art of 57.88\% normalized discounted cumulative gain. A generative visual dialog model equipped with the proposed technique also shows promising improvements.
\end{abstract}
\vspace{-10pt}
\section{Introduction}
\begin{figure}
	\centering
	\includegraphics[width=1.0\linewidth]{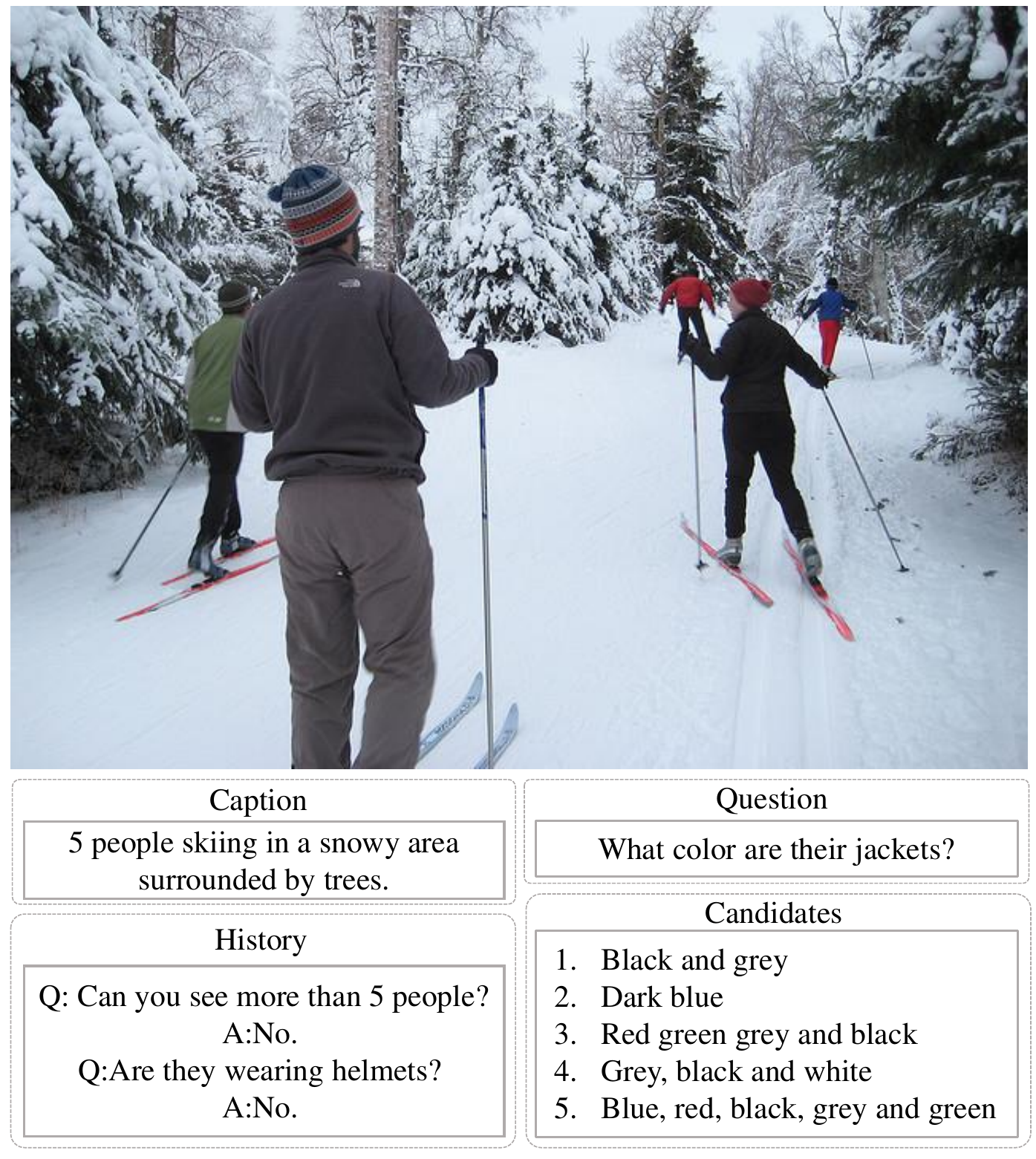}
	\caption{A general visual dialog task. The correct answer is picked from a candidate set by investigating the given image, caption, history, and question.}
	\label{fig:motivation1}
	\vspace{-5pt}
\end{figure}

Visual dialog is an emerging research theme lying at the intersection between computer vision and natural language processing. Given the capacities of reasoning, grounding, recognition, and translating, a visual dialog agent is expected to answer questions based on an image, caption, and history. For example, in Figure \ref{fig:motivation1}, the agent first reasons what the word `their' refers to in the current question based on history, locates the bounding boxes of the five people in the picture, recognizes the color of their jackets, and then translates the visual information into human language. Hence, a visual dialog task can also be regarded as: (\romannumeral1) visual grounding \cite{yu2018rethinking}, which further converts visual information in located bounding boxes into human language; (\romannumeral2) visual question answering (VQA) \cite{antol2015vqa}, which includes extra dialog history and caption as the input; and (\romannumeral3) image captioning \cite{vinyals2015show}, which generates a description not only based on visual information but also the history and question. 

A general visual dialog model has two components: an encoder to embed inputs (e.g., images and questions) into vectors and fuse them to create a unified representation, and a decoder either to translate the encoded vector directly into words for an answer or to rank the given candidate answers. Both VQA and visual dialog involve the fusion job of multiple modalities. However, as a multi-turn VQA task, in each turn, a visual dialog system must also integrate the caption and dialog history from past turns. Visual dialog systems can be grouped into two main categories according to different decoders: generative models and discriminative models. Generative models usually employ seq2seq \cite{sutskever2014sequence} or advanced reinforcement learning \cite{wu2017you} techniques to generate the answer set, where the highest probability answer is chosen as the output. Discriminative models tend to calculate the similarity between the latent output of the encoder and the embedding of candidate answers \cite{das2017visual, lu2017best, jain2018two, wu2017you}, with the correct answer expected to have the highest score. However, existing generative models aim to generate a single word of high probability at each step but omit the meaning of the whole answer sentence. The discriminative models are beneficial to understanding the answer sentence through long short-term memory (LSTM) \cite{hochreiter1997long}, but the scoring method is insufficient to capture the similarity between inputs and answers, since the vector of inputs and answers have been separately learned without deep fusion. Furthermore, both generative and discriminative models tend to give short and safe answer, such as `Yes' or `No', as their fusion methods focus on the major signal in short answer but will not look into details in a longer one. 

\begin{figure}
	\centering
	\includegraphics[width=1.0\linewidth]{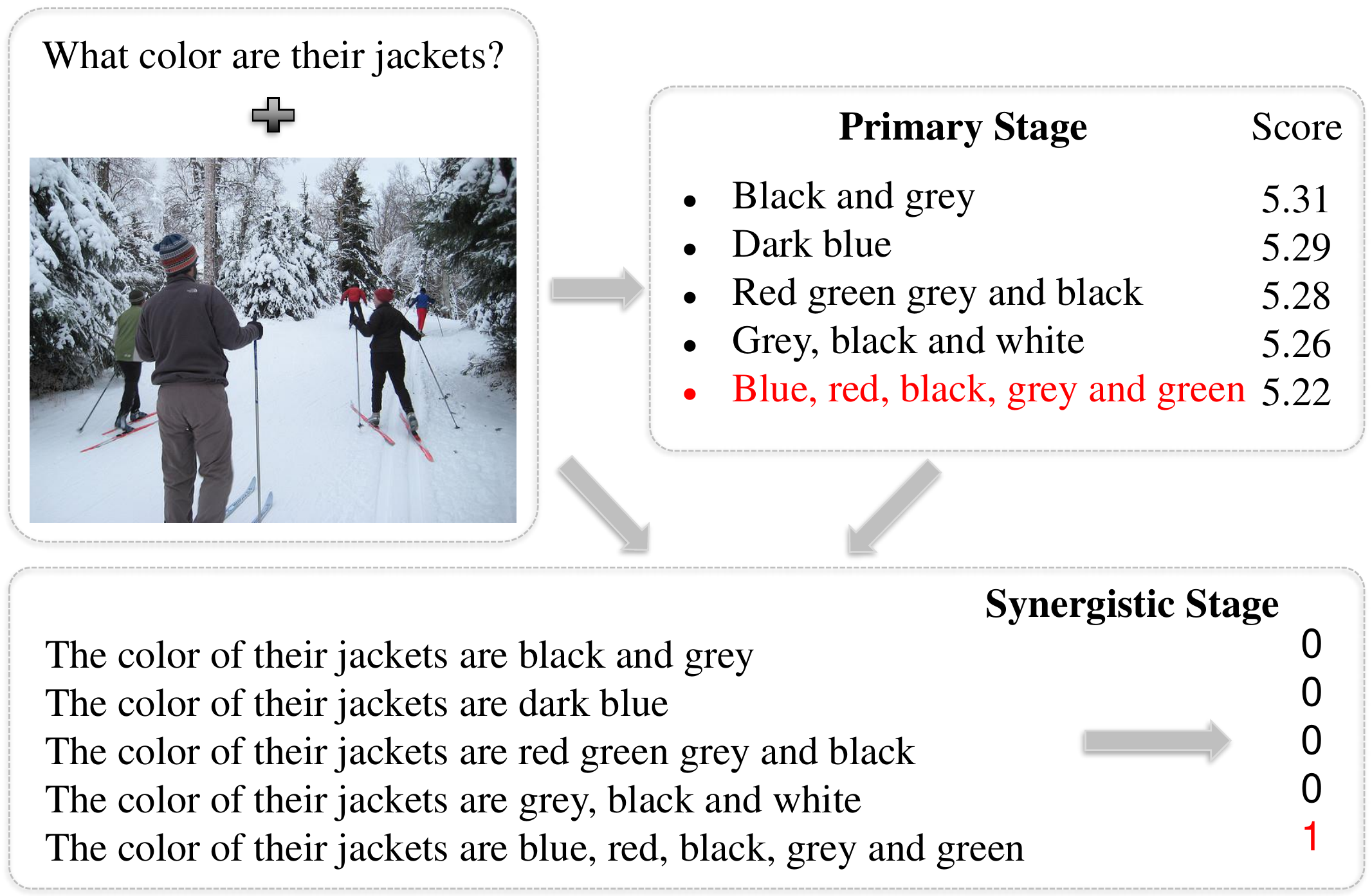}
	\caption{Candidate answers synergized with image and question are re-scored. The synergistic stage refers the answers back to the question and re-matches the image.}
	\label{fig:motivation2}
	\vspace{-5pt}
\end{figure}

To highlight the role of answer and its integration with other ingredients (e.g., images and questions) in visual dialog, we propose an image-question-answer synergistic network. However, not all answers are plausible for the image and question. For example, in Figure \ref{fig:motivation1}, only answers about color in the image are related to the question, while the others are unreasonable and probably lead to network side effects. Hence, we extend the traditional one-stage model composed of an encoder and a decoder into a two-stage model containing a primary stage and a synergistic stage. The primary stage can be any existing model that coarsely scores all candidate answers or generates some high probability candidates. The synergistic stage selects answers of high probability based on certain policies, finely synergizes them with the question and then re-ranks the candidate answers according to the relevance of the synergies to the image, as shown in Figure \ref{fig:motivation2}. The proposed method is consistent with human practice. In multiple choice examinations, we usually first exclude obviously wrong answers before paying more effort to compare the remaining answers that are more likely to be correct. We fill each answer into the question blank and judge whether the complete sentence is more suitable than the others. In addition, we address the class imbalance problem in the primary stage for a strong discriminative model. As a large number of easy negative candidate answers dominate the loss function, a temperature factor is considered in the loss function to discount the contribution of easy samples. 

Our model is evaluated on the Visual Dialog v1.0 dataset \cite{das2017visual}. In validation, our primary stage with a loss-balanced discriminative model improves the mean reciprocal rank (MRR) by 0.71\% compared to the non-balanced model, and the synergistic stage gives additional 0.91\% improvement on MRR. Furthermore, the synergistic stage in our generative model improves MRR by 4.7\% and recall on top-5 responses (R@5) by 9.2\% compared to the primary stage, which provides a different way to generate descriptive answers other than GAN and reinforcement learning \cite{wu2017you}. On the test-standard dataset, our two-stage model outperforms the baselines and achieves state-of-art performance, higher than the other entries in the Visual Dialog Challenge 2018 with a 57.88\% normalized discounted cumulative gain (NDCG).

\section{Related Work}
\textbf{Visual Question Answering (VQA)}: VQA is the first task undertaken when querying images for text answer generation. It is a classification problem in which candidate answers are restricted to the most common answers appearing in the dataset. Current models can be classified into three main categories: early fusion models, later fusion models, and external knowledge-based models. In early fusion models, the input queries are regarded as parameters of conditional batch normalization \cite{de2017modulating, perez2017learning} in the detection network \cite{he2016deep}, which leads a pre-trained ResNet to the proposed MODERN architecture; this method affects less than 1\% of parameters in the pre-trained model, which reduces the risk of over-fitting. Later fusion models mainly concentrate on how to represent the answer vector by jointing the question and global image feature \cite{jabri2016revisiting, antol2015vqa}. However, a lot of visual information is irrelevant to the input query, resulting in significant noise during reasoning. Thus, the attention mechanism is introduced to solve this problem. This starts with the linear combination of two features, such as the stacked attention network \cite{yang2016stacked} by learning the attention using multi-step reasoning and the dual attention network \cite{nam2016dual} that learns both the visual and textual attention. Then, a bilinear pooling method like multi-modal compact bilinear pooling (MCB) \cite{fukui2016multimodal} is applied by projecting the outer product of two features into the high dimension of quadratic expansion. However, MCB needs to sample features, which is computationally intractable and has very large projected dimension, so low-rank bilinear pooling (MLB) \cite{kim2016hadamard} and multi-factor bilinear pooling (MFB) \cite{yu2017mfb} have been proposed to project the two features into a common low-rank space, and the bilinear attention network (BAN) \cite{kim2018bilinear} builds the interaction attention between multi-modal inputs. In contrast to early and later fusion models, external knowledge-based models suppose that common sense or information not given in the image is required to infer the right answer. \cite{wu2016ask} uses DBpedia \cite{auer2007dbpedia} to broaden the range of answers. \cite{narasimhan2018straight} queries the triplet (visual concept, relation, attribute) in Fvqa \cite {wang2017fvqa} to score the retrieved facts. However, this method is not good enough to reason complex facts and requires further development. 

\textbf{Visual Dialog}: Extending the single turn dialog task (VQA) to a multi-turn one, we introduce visual dialog. Before the dataset proposed by \cite{das2017visual}, \cite{de2017guesswhat} used a dataset that located the object in an image in which one person gave a question about the image and the other provided a `Yes/No/NA' answer based on the truth. The current dataset expands the range of questions and answers. Question types about image can be various including color, number, relationships, etc., while the answer could be simple as a `No' or a complex description of the image. Three encoding methods were provided in \cite{das2017visual} as baselines, namely late fusion, the hierarchical recurrent encoder, and the memory network, as well as two decoding methods: LSTM and softmax. Inspired by the generative adversarial networks (GANs) \cite{goodfellow2014generative} and the performance gap between discriminative and generative networks, \cite{lu2017best} transferred knowledge from a pre-trained discriminative network to a generative network with a Gumbel-softmax \cite{jang2016categorical} LSTM encoder using perceptual loss. \cite{wu2017you} combined GAN and reinforcement learning \cite{yu2017seqgan} to train the generator with a co-attention encoder, thereby allowing the discriminator to directly access the generated response to evaluate its quality, and used Monte Carlo (MC) searching with a roll-out policy to compute the intermediate reward for each word. \cite{jain2018two} merged the fusion step and the scoring step into a single step, which is similar to our synergistic stage, but involved all the image information containing noise and the whole history containing topics irrelevant to current question. Furthermore, it simply arrayed the isolated vectors of the image, question, history, and answer for their representation.    

\section{Synergistic Network}
\begin{figure*}[t!]
	\centering
	\includegraphics[width=1.0\linewidth]{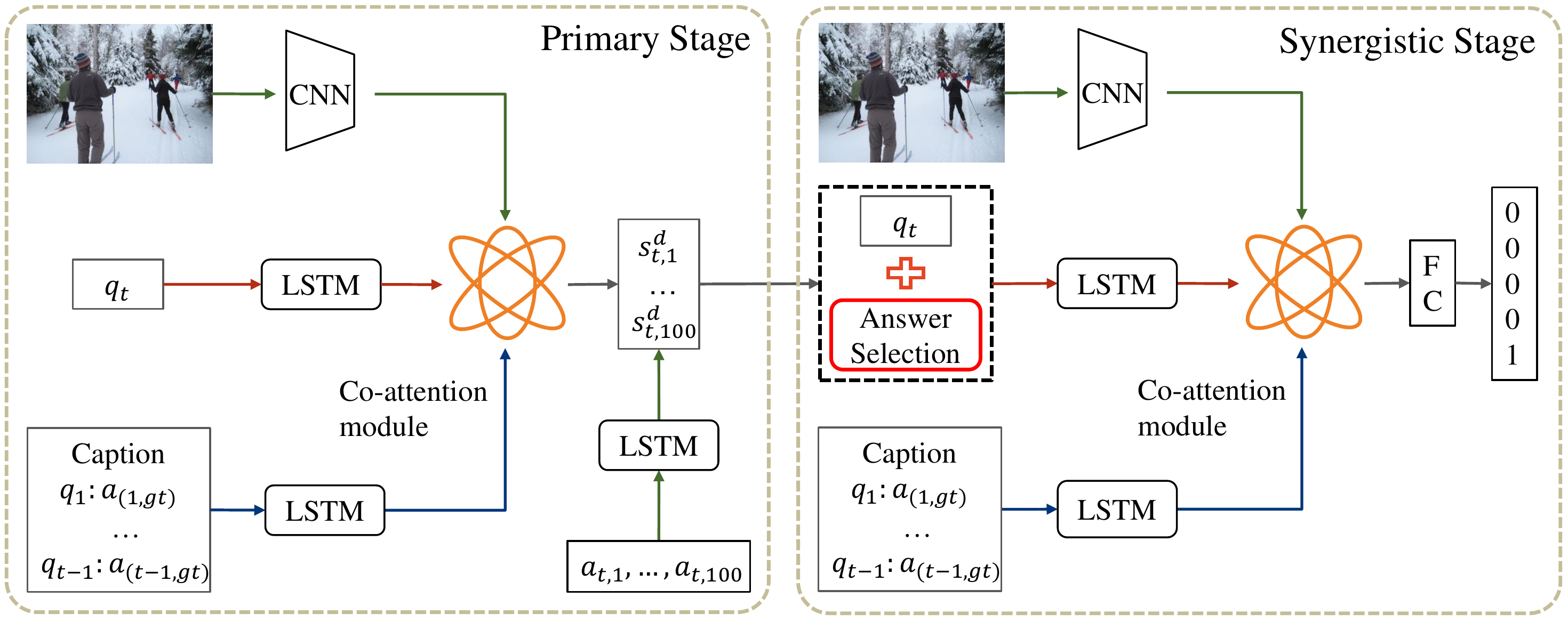}
	\caption{Architecture of our model. All candidate answers are scored in the primary stage, and some selected answers are re-scored in the synergistic stage.}
	\label{fig:framework}
	\vspace{-5pt}
\end{figure*}

In a candidate answer set, some of the answers are hard samples that are close or equal to the correct answer, while others are easy samples. Our framework shown in Figure \ref{fig:framework} has two stages, the primary stage and the synergistic stage. In the primary stage, we learn representative vectors of the image, history, and question using a co-attention module and then calculates the score of each candidate answer to separate hard answers from easy ones. In the synergistic stage, we select hard answers together with their questions to form question-answer pairs. These pairs further coordinate with the image and history to predict scores. 

We first formally define the visual dialog problem and then introduce our new synergistic strategy. Given an image $I$ and caption $C$, we collect historical questions and their corresponding answers as $H$. At turn $t$, our model gives a score for each answer $a_{t,i}$ in candidate set $A_t$ based on question $q_t$. To describe input image $I$, we detect objects and their features using the Faster-RCNN model \cite{anderson2018bottom} and apply CNN to encode them into $V = (v_1, \dots, v_n)$, where $v_i \in R^d$ and $n$ is the number of objects. Question $q_t$ is a sequence of words, which can be encoded using LSTM, i.e., $m^q_t = \text{LSTM}(q_t)$. We also organize the previous dialog (including caption) as history $H = (H_0, \dots, H_{t-1})$, where $H_0 = C$, and $H_i = (q_i, a_{i,gt})$ for $i \in \{1, \dots, t-1 \}$, which is the concatenation of the question and correct answer at each turn before time $t$. Similar to the question, we use another LSTM to extract the history features as $U = (u_0, \dots, u_{t-1})$, where $u_i = \text{LSTM}(H_i)$. $m^q_t \in R^d$ and $u_i \in R^d$ correspond to the last state of LSTM.

\subsection{Primary Stage}\label{primary}
An encoder-decoder solution is adopted in the primary stage \cite{das2017visual, wu2017you, lu2017best}. The encoder contains two main tasks, one is how to de-reference in the multi-turn conversations (98\% of dialogs contain at least one pronoun), and the other is to locate the objects in the image mentioned in the current question. The attention mechanism \cite{lu2017best} is commonly used to tackle the tasks. Instead of linear concatenation, we use multi-modal factorized bilinear pooling (MFB) \cite{yu2017mfb}, as it can overcome the difference between distributions of the two features (two LSTMs to encode question and history, respectively; LSTMs for text feature and CNNs for image feature). MFB is expected to provide a richer representation than other bilinear methods, such as MLB \cite{kim2016hadamard} and MCB \cite{fukui2016multimodal}. In MFB, the fusion of two features $X, Y \in R^d$ is calculated by:
\vspace{-10pt}
\begin{equation}
z = \mathrm{MFB}(X, Y) = \sum_{i=1}^{k}(\mathbf{U}^\top_iX \circ \mathbf{V}^\top_iY),
\vspace{-10pt}
\end{equation}
where $\mathbf{U}, \mathbf{V} \in R^{d \times l \times k}$ are the parameters to be learned, $k$ is the number of factors, $l$ is the hidden size, and $\circ$ is the Hadamard product (element-wise multiplication). However, $Y$ sometimes represents multiple channel input, e.g., detected objects or history in our model, so the formula becomes:
\vspace{-10pt}
\begin{equation}
z = \mathrm{MFB}(X, Y) = \sum_{i=1}^{k}((\mathbf{U}^\top_iX \cdot \mathds{1}^\top)  \circ (\mathbf{V}^\top_iY)),
\vspace{-10pt}
\end{equation}
where $\mathds{1} \in R^\phi$ is the vector with all elements equal to one, and $\phi$ is the channel number of $Y$. To stabilize the output neurons, we use power normalization ($z \leftarrow \mathrm{sign}(z)|z|^{0.5}$) and $\ell_2$ normalization ($z \leftarrow z/\|z\|$). 

We utilize MFB to learn the unified vector of the question and history, denoted as $z^h_t = \mathrm{MFB}_h(m^q_t, U)$, where $z^h_t \in R^{l \times t}$. Then, we learn the attention weight and vector by:
\vspace{-10pt}
\begin{eqnarray}
\label{eq:attention1}
\alpha^h_t &=& \text{softmax}(w^\top_\alpha z^h_t), \\
\label{eq:attention2}
m^h_t &=& \sum_{i=0}^{t-1}(\alpha^h_{t,i}u_i),
\end{eqnarray}
$w^\top_\alpha \in R^l$ is a learned parameter, and $\alpha^h_t \in R^t$ is the calculated weight implying which history the question should refer to. The attended history vector $m^h_t$ can be concatenated with the question vector and then fused with image features, $z^v_t = \mathrm{MFB}_v([m^q_t:m^h_t], V)$. The image attention vector $m^v_t$ can be obtained in a similar approach as Eq.(\ref{eq:attention1}) and Eq.(\ref{eq:attention2}) by taking $z^v_t$ as the input. Finally, we learn the representation of text and visual features with $e_t^p = \mathrm{MFB}_e([m^q_t:m^h_t], m^v_t)$.

The decoder encodes each candidate answer $a_{t, i} \in A_t$ to $m^a_{t,i}$ using LSTM and calculates the dot similarity score by:
\vspace{-5pt}
\begin{equation}
s^d_{t, i} = e_t^p\top f_d(m^a_{t,i}),
\end{equation}
where $f_d$ is a one-layer MLP with activation $tanh$ to project the answer encoding $m^a_{t,i} \in R^d$ to the space of input embedding $e^p_t$. 

The correct answer $a_{t,gt}$ should have a higher score than the others. Thus, we use N-pair loss \cite{sohn2016improved} to measure the error. Most of the 100 candidate answers are easy samples, which are irrelevant to the inputs, and contribute to no useful learning signal in this loss (score difference less than zero in Figure \ref{fig:loss}). To solve the imbalance problem, we employ temperature $\tau$ to reduce the imbalance impact:
\vspace{-5pt}
\begin{equation}
\mathcal{L}_D = \log (\sum_{i=1}^{100}\exp \dfrac{s^d_{t, i} - s^d_{t, gt}}{\tau}),
\vspace{-5pt}
\end{equation}
where $\tau \leq 1$. If the candidate answer $a_{t, i}$ is correctly scored lower than that of the ground truth answer $a_{t, gt}$, loss $l_{t, i} = s^d_{t, i} - s^d_{t, gt}$ will be less than 0, and $\tau$ can reduce the contribution of answer $a_{t,i}$. For instance, with $\tau = 0.25$ and $l_{t, i} = -1$, about 20 such items make the same loss as the normal N-pair loss. Otherwise, it amplifies the loss of incorrectly scored answers.

\subsection{Synergistic Stage}\label{refine}
In the primary stage, some answers are improperly scored due to the limitations of the scoring method. In this synergistic stage, answers are coordinated with the question and image for re-rank. However, easy candidate answers are not needed in further analysis, and we want our second stage model to fully focus on hard answers within its modeling capacity. Therefore, we select the answers with higher probability of being correct based on the predicted scores from the primary stage. As seen from the recall of our best method in Table \ref{table:challenge 2018}, the top ten answers predicted in the primary stage covered nearly 90\% of the ground truth, which means that the remaining 90 answers are of lower probability and can easily be discriminated. Based on this phenomenon, we first pick the top $N$ answers from $A_t$ to organize a new candidate set $B_t$, where $B_t = (b_{t,1}, \dots, b_{t,N}), B_t \subset A_t$. 

The selected answers are often ambiguous for describing the whole meaning of sentences (such as `No' and `Black and grey' in Figure \ref{fig:motivation1}), so they must work with the corresponding question to make complete sentences. Thus, we append question $q_t$ to each answer $b_{t,j}, j \in \{1, \dots, N\}$ as a question-answer pair and encode it using LSTM to obtain a vector:
\begin{equation}
m^b_{t, j} = \text{LSTM}([q_t:b_{t,j}]).
\end{equation}
Extra history is required to remedy the reference problem of the question. Therefore, we use $m^b_{t,j}$ as a question vector combined with attended history $m^h_t$ to learn the image's attention parameters:
\vspace{-5pt}
\begin{equation}
z^{r}_{t,j} = \mathrm{MFB}_a([m^b_{t, j}:m^h_t], V)
\vspace{-5pt}
\end{equation}
and the attended image feature $m_{t,j}^r$ using Eq.(\ref{eq:attention1}) and Eq.(\ref{eq:attention2}) for selected answer $b_{t,j}$. Similar to the primary stage, we obtain the fusion embedding $e_{t,j}^r = \mathrm{MFB}_r([m^b_{t, j}:m^h_t], m_{t,j}^r)$, which represents the answer vector synergized with the image, question, and history, which is directly used to calculate the score by:
\vspace{-5pt}
\begin{equation}
s^r_{t, j} = f_r(e_{t,j}^r),
\vspace{-5pt}
\end{equation}
where $f_r$ can be a one-layer MLP. An answer in candidate set containing more details and better matching the inputs should earn a higher score than ordinary ones.

Here, we reuse the attended history vector $m^h_t$ from the primary stage, since the co-reference problem in the question can be resolved without knowing the answer. For example, as shown in Figure \ref{fig:motivation1}, we can refer `their' to the `five people' regardless of the colors of their jackets. The image feature $V$ is shared with the primary stage, because we want the image feature to be represented universally in the two stages. But the attention weights are learned at each stage, since each candidate answer, as well as the question, depicts its own attention map. 

We treat this stage as a classification problem, where the correct answer should have the highest probability:
\vspace{-5pt}
\begin{eqnarray}
p^r_t &=& \text{softmax}(s^r_t),\\
\mathcal{L}_R &=& \sum_{j=1}^{N}-y_j\log(p^r_{t,j}),
\end{eqnarray}
where $y_{gt}$ is equal to 1 and the others are zeros. We note that this formula can be easily extended to soft cross entropy, where $y_i$ is the probability marking this answer as correct if a dense annotation dataset is available in the future.

\section{Extension to the Generative Model}
\begin{figure}
	\centering
	\includegraphics[width=1.0\linewidth]{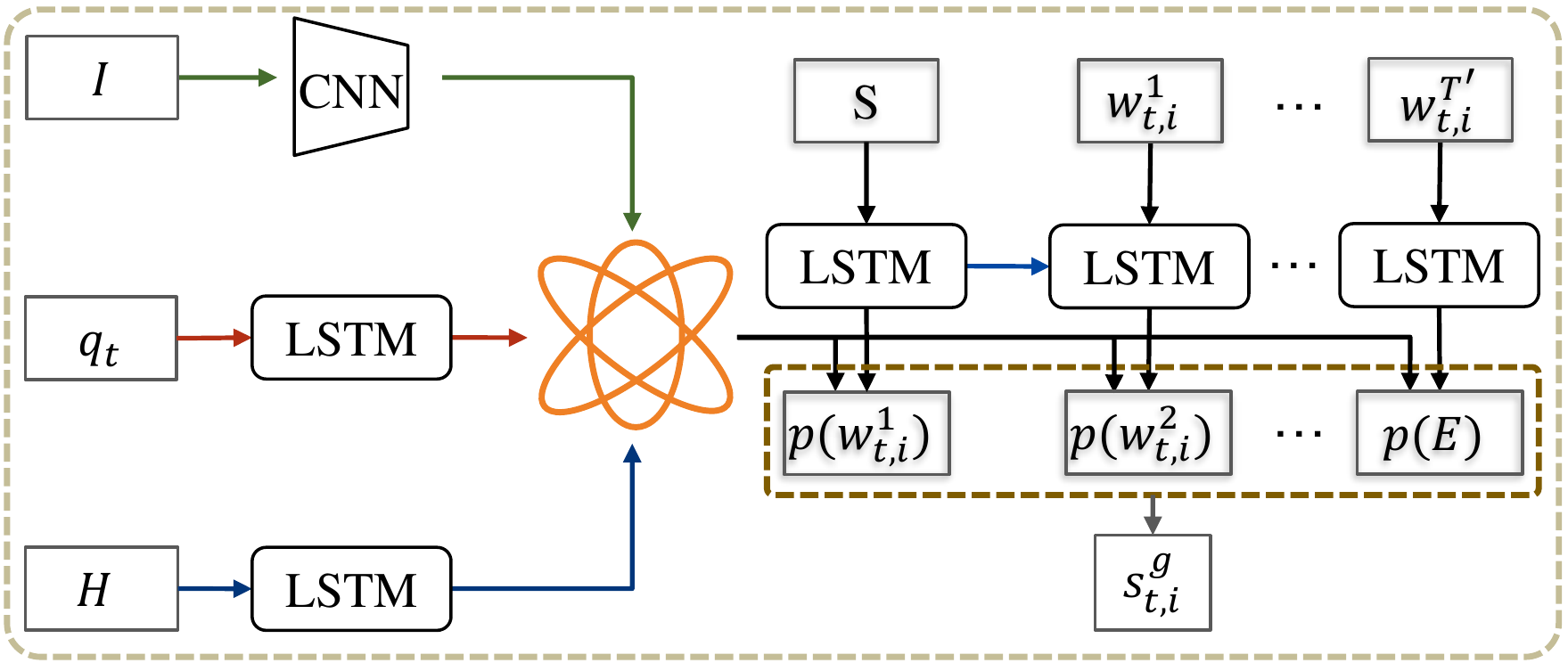}
	\caption{Primary stage of our generative model. The score of each answer is its probability of word sequence.}
	\label{fig:framework1}
	\vspace{-5pt}
\end{figure}

Besides discriminative model in the primary stage, the generative model can also be used to score the candidate answers and seamlessly works with the proposed image-question-answer synergistic method. The encoder of the generative model is the same as that of the discriminative model in the primary stage in Section \ref{primary}. Accordingly, we still use $e^p_t$ to represent the common vector of image $I$, history $H$, and question $q_t$ at turn $t$. The decoder interprets $e^p_t$ into answers and calculates the probability of answer $a_{t, i}$ by:
\vspace{-15pt}
\begin{equation}
\label{eq:gen1}
\begin{split}
s^g_{t, i} &= p(w^1_{t,i}, \dots, w^{T'}_{t,i}|q_t, H, I) \\
&= \prod_{j=1}^{T'}p(w^j_{t, i}|(w^1_{t,i}, \dots, w^{j-1}_{t,i}), e^p_t),\\
\end{split}
\end{equation}
which is also regarded as score in the primary stage as shown in Figure \ref{fig:framework1},
where $w^1_{t, i}, \dots, w^{T'}_{t,i}$ is the word sequence of the answer $a_{t, i}$, and $T'$ is word number. For each word, its probability is given by:
\begin{equation}
p(w^j_{t, i}|(w^1_{t,i}, \dots, w^{j-1}_{t,i}), e^p_t) = f_g(\mathrm{MFB}_g(h_j, e^p_t)),
\end{equation}
where $h_0$ is the last hidden state of LSTM for the question, $h_j=\text{LSTM}(h_{j-1}, w^{j-1}_{t,i})$ is the state of LSTM, and $f_g$ maps the fusion vector to the word space. Instead of initializing the LSTM of decoder with encoded vector $e_t^p$ \cite{das2017visual}, we regard it as a context vector \cite{lu2017knowing}. This is for three main reasons: first, the encoded vector and the LSTM decoder have different distributions, and all gates and the hidden state in LSTM are learned by linear combination; second, the salient objects that should be attended by each token are already chosen by the question, and the aim of the decoder is to translate the visual information of salient objects into text, such that the context vector is fixed for each token; and third, instead of learning a joint vector with $h_{j-1}$, the context vector could be considered as compensatory information to the current hidden state $h_j$, which reduces the uncertainty for next word prediction. 

To make the correct answer $a_{t, gt}$ score higher in the primary stage, we maximize the conditional probability $p(w^1_{t, gt}, \dots, w^{T'}_{t,gt}|q_t, H, I)$. Thus, the loss function is the sum of the negative log likelihood of the correct word in each step:
\vspace{-5pt}
\begin{equation}
\mathcal{L}_G = -\sum_{i=1}^{T'} \log p(a_{t, gt}).
\vspace{-5pt}
\end{equation}

If the candidate answers are given in the generative model as in the discriminative model, we can collect the score for each answer using Eq.(\ref{eq:gen1}). Otherwise, we can generate some candidate answers with high probability by beam search \cite{sutskever2014sequence}. After collecting the score for each answer in the primary stage, we follow the strategy in Section \ref{refine} to pick the top $N$ answers and then synergize them with the image, question, and history to learn a better representation for re-scoring.

\vspace{-5pt}
\section{Experiments}
In this section, we evaluate our synergistic strategy on a visual dialog dataset. We introduce the dataset and evaluation metric, before describing our experimental setting and results, and finally the qualitative analysis.

\vspace{-5pt}
\subsection{Dataset and Evaluation Metric}
Our model is trained on the Visual Dialog v1.0 dataset \cite{das2017visual}, which contains about 120k images from COCO-trainval \cite{lin2014microsoft}. Each image has one caption and 10 turn dialogs, i.e., about 1.2 million question-answer pairs. To organize this dataset, two people chatted on the Amazon Mechanical Turk. The questioner could not see the image and asked a question based on the given caption and previous context to better understand the scene, while the answerer could see both the image and caption and replied to the question as naturally and conversationally as possible. Each question has 100 candidate answers containing one correct answer, 50 answers to similar questions, 30 popular answers, and some randomly picked answers from the dataset. For the validation and test datasets, 10K COCO-like images were collected from Flickr. The test dataset was densely annotated in v1.0 by four people to allow application for the more robust evaluation metric NDCG \cite{wang2013theoretical} rather than traditional retrieval metrics, such as mean rank, R@1,5,10, and MRR. There could be more than one correct answer to each question in the candidate set, such as `yeah' and `yes'. In this situation, NDCG is invariant to the order of options with identical relevance. For each candidate answer, its relevance is $\frac{\text{annotators who marked answer as relevent}}{4}$. The metric is given by:
\vspace{-10pt}
\begin{eqnarray}
\text{DCG@k} &=& \sum_{i=1}^{\text{k}}\frac{\text{relevance}_i}{\log_2(i+1)},\\
\text{NDCG@k} &=& \frac{\text{DCG@k for submited ranking}}{\text{DCG@k for ideal ranking}},
\end{eqnarray}
where k is the number of answer options whose relevance is greater than zero. Of these metrics, a higher score is better for NDCG, MRR, and R@1,5,10, but a lower score is better for mean rank.

\vspace{-5pt}
\subsection{Implementation Details}
We first constructed the vocabulary, which contains the words appearing in the questions, correct answers, and captions more than four times in the train dataset. This made 11,213 words with the padding word `PAD', out-of-vocabulary word `UNK', start symbol `START', and end symbol `END'. Then, each word was embedded within a 300-dimension vector shared across caption, history, question, and answer. The maximum lengths of the caption, question, answer, and history were 40, 20, 20, and 40, respectively. For each candidate answer, we inserted `START' at the head and appended `END' at the tail. The LSTMs of the question and history are two layered, while it is one layered for the answer in the primary stage and the question-answer pair in the synergistic stage. The hidden state dimension $d$ for all LSTMs and CNN is 512. For bilinear pooling, we set $k$ to 5 and $l$ to 1000 as Yu \cite{yu2017mfb}.  

We start training the primary stage with loss $\mathcal{L}_D$ ($\mathcal{L}_G$) for 7 epochs to arrange the top ranked answers relative to the inputs. The synergistic stage follows with loss $\mathcal{L}_D (\mathcal{L}_G) + \mathcal{L}_R$ for another 15 epochs. When training the synergistic stage, our policy is to randomly sample $N-1$ answers from the top $M$ ranked in the first stage combined with the right answer $a_{t, gt}$ to organize $B_t$. During testing, we only select the top $N$ answers. For this dataset, we choose $N = 10 \text{ or } 15$ and $M$ varying from 10 to 40 for the discriminative model, then $N = 10, 20, 30$ and $M$ fixed to 30 for the generative model, and we analyze the effect of different $N$ and $M$ on performance in Section \ref{ablation}. Our model is trained using the Adam solver \cite{kingma2014adam} with $\beta_1=0.9$, $\beta_2=0.99$, initial learning rate $10^{-3}$, and decay every 7 epochs with an exponential rate of 0.25.

\vspace{-5pt}
\subsection{Comparison with the State-of-the-art}
We compare our discriminative model with baselines \cite{das2017visual} and other methods: Later Fusion (\textbf{LF}), which encodes the question, image, and history respectively and projects their concatenation into a joint embedding; Hierarchical Recurrent Encoder (\textbf{HRE}), which uses a hierarchical architecture to encode the dialog history; Memory Network (\textbf{MN}), which maintains a memory bank storing previous dialog, which is attended by a question; and \textbf{MN-att} and \textbf{LF-att}, which add an attention mechanism for image to their base methods. From Table \ref{table:challenge 2018}, it can be seen that our best single model improves NDCG by 7.56\% and MRR by 5.13\% compared with \textbf{LF-att}. To improve accuracy, we ensemble 10 models with different seeds and $M$. We rank the answers by summing scores of all models and achieve the highest NDCG on the test-standard server of Visual Dialog Challenge 2018.

\begin{table}
	\centering	
	\resizebox{\linewidth}{!}{
	\begin{tabular}{lccccccc}
		\hline
		\textbf{Model} & \textbf{NDCG} & \textbf{MRR} & \textbf{R@1} & \textbf{R@5} & \textbf{R@10} & \textbf{Mean} \\
		\hline
		LF\cite{das2017visual} & 45.31 & 55.42 & 40.95 & 72.45 & 82.83 & 5.95 \\
		HRE\cite{das2017visual} & 45.46 & 54.16 & 39.93 & 70.45 & 81.50 & 6.41 \\
		MN\cite{das2017visual} & 47.50 & 55.49 & 40.98 & 72.30 & 83.30 & 5.92 \\
		MN-att\cite{das2017visual} & 49.58 & 56.90 & 42.43 & 74.00 & 84.35 & 5.59 \\
		LF-att\cite{das2017visual} & 49.76 & 57.07 & 42.08 & 74.83 & 85.05 & 5.41 \\
		Technion & 54.46 & 67.25 & 53.40 & 85.28 & 92.70 & 3.55 \\
		MS AI & 55.35 & 63.27 & 49.53 & 80.40 & 89.60 & 4.15 \\
		USTC-YTH & 56.47 & 61.44 & 47.65 & 78.13 & 87.88 & 4.65\\
		\hline
		Single (ours) & 57.32 & 62.20 & 47.90 & 80.43 & 89.95 & 4.17 \\
		Ensemble (ours) & 57.88 & 63.42 & 49.30 & 80.77 & 90.68 & 3.97 \\
		\hline\\
	\end{tabular}}
	\vspace{-10pt}
	\caption{Performance of discriminative models on test-standard server of Visual Dialog Challenge 2018. We ensemble 10 models with different seeds and varying $M$ from 25 to 35.}
	\vspace{-10pt}
	\label{table:challenge 2018}
\end{table}

\vspace{-5pt}
\subsection{Ablation Study}\label{ablation}
\begin{figure}
	\centering
	\includegraphics[width=1.0\linewidth]{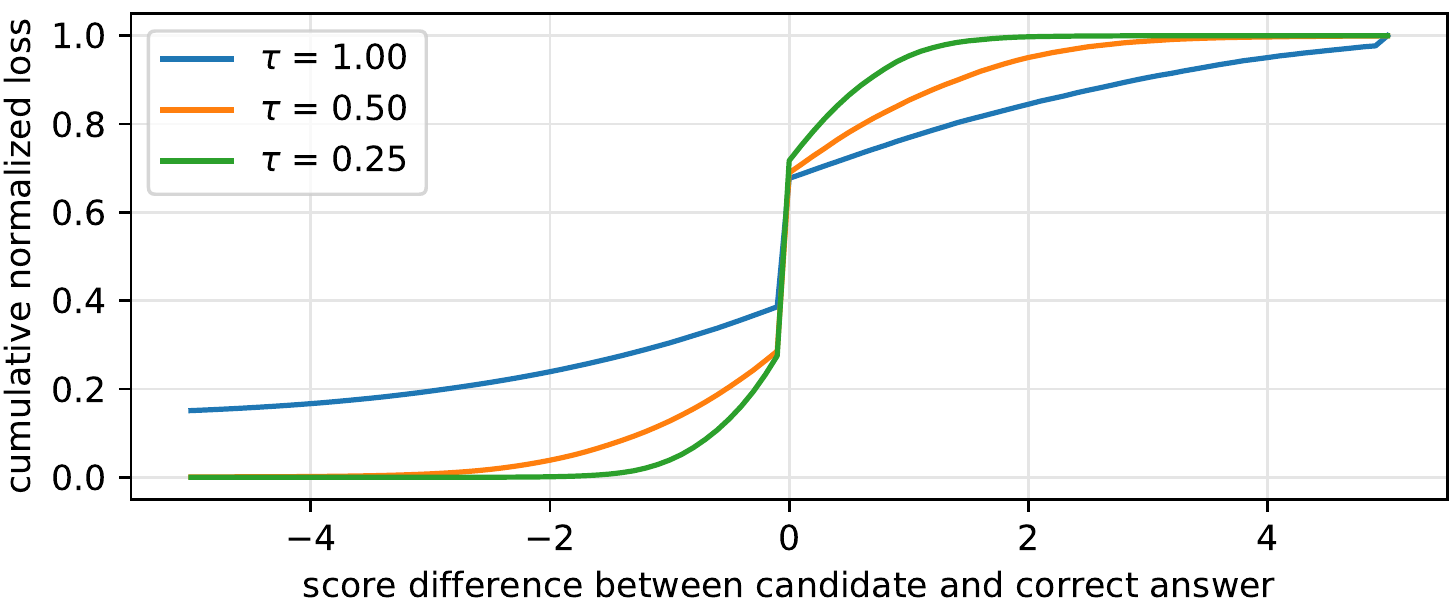}
	\vspace{-10pt}
	\caption{Cumulative normalized loss for different $\tau$.}
	\label{fig:loss}
	\vspace{-10pt}
\end{figure}

We conduct several ablation studies to verify the contribution of each module to our discriminative model. The first three lines in Table \ref{table:dis_model} show the performance of the primary stage with different $\tau$ varying from 1 to 0.25. The performance improves as $\tau$ decays, because most candidate answers are easily discriminated. The summed loss of these easy negative answers whose scores are lower than the correct answer by one consumes almost 30\% of the model's energy at $\tau=1.00$, as shown in Figure \ref{fig:loss}. This kind of loss is reduced as $\tau$ decreases and is nearly zero when $\tau=0.25$ , saturating our primary stage. This makes the model pay more attention to incorrect answers scored near or higher than the correct answer. The last five lines show the results of the synergistic stage with different settings. MRR of this stage drops at $N=10$ and $M=10$, as the top answers of the primary stage become stable after several epochs, resulting in the synergistic stage learning bias. Feeding more samples by increasing $M$ improves performance to produce the best model at $M=30$, showing that synergy can learn a better representation of the image, question, and answer. What of interest is that the performance drops at $M=40$ or $N=15$, possibly because the selected answers become less relevant to the inputs, and the second stage model's ability is to score nearly correct answers but is sensitive to unrelated ones. Conversely, it also addresses the importance of the primary stage to the synergistic stage. And the primary stage is also necessary to balance the memory cost, since each answer learns its own attention map and fusion with the image in the synergistic stage, while in the primary stage, only one attention map is required for the question and image.
 
\begin{table}
	\centering	
	\resizebox{\linewidth}{!}{
		\begin{tabular}{cccccccc}
			\hline
			\textbf{N} & \textbf{M} & $\tau$ & \textbf{MRR} & \textbf{R@1} & \textbf{R@5} & \textbf{R@10} & \textbf{Mean}\\
			\hline
			- & - & 1.00 & 61.92 & 47.53 & 79.78 & 89.28 & 4.42 \\
			- & - & 0.50 & 62.51 & 48.30 & 80.05 & 89.57 & 4.27 \\
			- & - & 0.25 & 62.63 & 48.31 & 80.39 & 89.65 & 4.23 \\
			10 & 10 & 0.25 & 62.31 & 48.18 & 79.45 & 90.21 & 4.22 \\
			10 & 20 & 0.25 & 62.83 & 48.45 & 80.70 & 90.28 & 4.11 \\ 
			10 & 30 & 0.25 & \textbf{63.54} & \textbf{49.21} & \textbf{81.01} & \textbf{90.32} & \textbf{4.09} \\
			10 & 40 & 0.25 & 63.14 & 48.77 & 80.97 & 90.02 & 4.13 \\
			15 & 30 & 0.25 & 63.16 & 48.91 & 80.75 & 90.22 & 4.12 \\
			\hline\\
		\end{tabular}}
		\vspace{-10pt}
		\caption{Performance of our discriminative model on validation dataset.}
		\vspace{-5pt}
		\label{table:dis_model}
\end{table}

For the generative model, the first line in Table \ref{table:gen_model1} shows the performance of the memory network initialized by the common vector of inputs, and the second line shows the result of our model using only the primary stage, leaving other three lines for the synergistic stage. It can be seen that the primary stage outperforms baseline by 1.1\% with respect to MRR, which generates a strong candidate set for the next stage. Our synergistic stage further improves MRR by 2.6\% when $N=10$, since the primary stage focuses on each word but lacks the understanding of whole answer sequence. In contrast to the discriminative model in which about 90\% of correct answers are top ten ranked in the primary stage, only 2 out of 3 correct answers are top sorted by the generative model, so we increase the selected answer number $N$ from 10 to 30 in the second stage, which further increases MRR by 2.1\%. Furthermore, our model improves R@5 by 9.2\% and R@10 by 8.9\%.
\begin{table}
	\centering	
	\resizebox{\linewidth}{!}{
		\begin{tabular}{lccccccc}
			\hline
			\textbf{Model} & \textbf{N} & \textbf{M} & \textbf{MRR} & \textbf{R@1} & \textbf{R@5} & \textbf{R@10} & \textbf{Mean}\\
			\hline
			MN-att \cite{das2017visual} & - & - & 47.94 & 37.48 & 58.56 & 65.57 & 17.61 \\
			Primary (ours) & - & - & 49.01 & 38.54 & 59.82 & 66.94 & 16.69 \\
			Synergistic (ours) & 10 & 30 & 51.62 & 40.77 & 63.58 & 67.00 & 16.51 \\
			Synergistic (ours) & 20 & 30 & 53.23 & \textbf{41.42} & 67.22 & 72.91 & 15.87 \\
			Synergistic (ours) & 30 & 30 & \textbf{53.73} & 41.28 & \textbf{69.01} & \textbf{75.85} & \textbf{15.12} \\
			\hline\\
		\end{tabular}}
		\vspace{-10pt}
		\caption{Performance of our generative model on validation dataset with candidate set.}
		\vspace{-5pt}
		\label{table:gen_model1}
\end{table}

\vspace{-5pt}
\subsection{Qualitative Analysis}
\begin{figure*}
	\centering
	\includegraphics[width=1.0\linewidth]{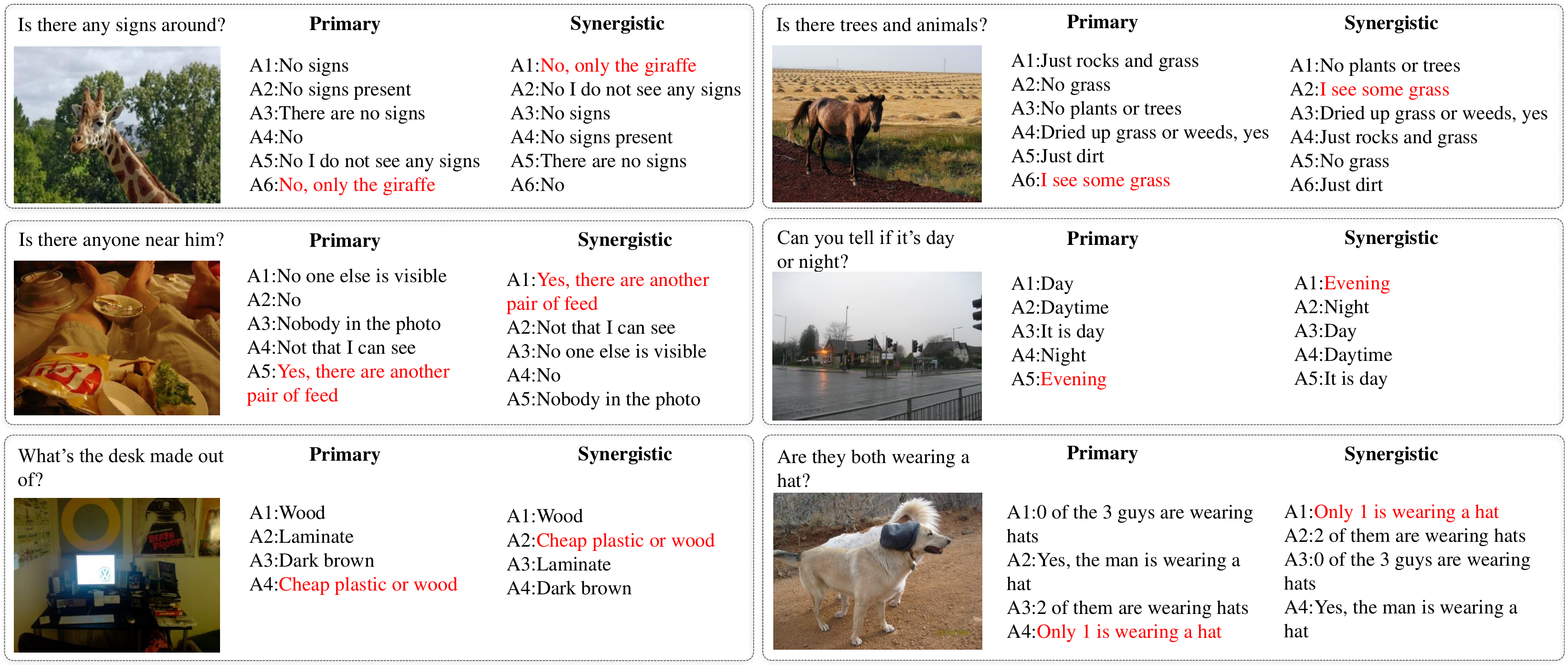}
	\caption{Qualitative comparison for discriminative model.}
	\label{fig:visualization1}
	\vspace{-10pt}
\end{figure*}
\begin{figure*}
	\centering
	\includegraphics[width=1.0\linewidth]{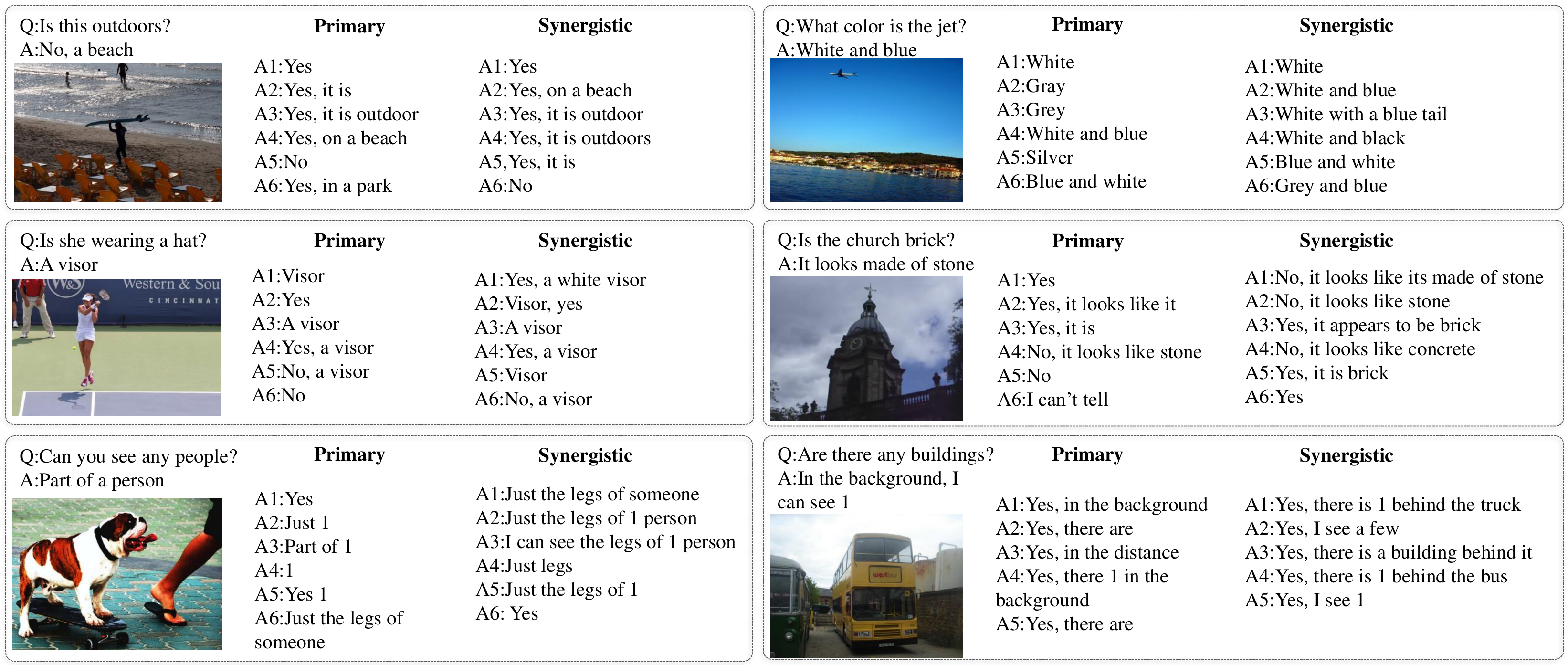}
	\caption{Qualitative comparison for generative model without candidate answers.}
	\label{fig:visualization2}
	\vspace{-10pt}
\end{figure*}

To further demonstrate the effects of our synergistic model, we present some examples from the validation dataset. Figure \ref{fig:visualization1} shows the results of the discriminative model ranked by only the primary stage and our two-stage network. The answer in red is correct, while the others are top-ranked candidate answers. It can be seen that the one-stage model tries to give a safe answer such as `No' to the top-left and middle-left images, while our model depicts more details by adding `only the giraffe' and `another pair of feed'. Biased answers are given in middle-right image, since for a binary question with choices, the answer word is always contained in the question. There is also bias in bottom-right image, because only humans tend to wear hats. Our model gives unbiased answers based on the images, and furthermore, can detect the discrepancies between similar words in the bottom-left image. 

In order to apply our method to realistic applications, we abandon the prepared candidate set in the first stage and generate another within our primary model by beam search \cite{sutskever2014sequence}, which maintains a partial sequence list of size $B$. At each step, all partial sequences are extended with the whole vocabulary, and only the top $B$ sequences with the highest probability are retained for next step. Sentences meeting the `END' symbol are moved from the partial list to the complete one. Starting with the `START' symbol and iterating for a maximum 20 steps, we obtain a candidate set of size $B$ with all complete sentences complemented by some partial ones. In Figure \ref{fig:visualization2}, we show the top generated answers with $B=15$ for the primary and synergistic stage, the preset answer is below the question. The primary stage always ranks short answers having one or two words higher than the long sequences that depict images with more information, because the generative method calculates the score for each answer by product probability of its words. In the synergistic stage, this problem is overcome, since the extra attribute information, e.g., `a white visor' (middle-left image) and `white with a blue tail' (top-right image), can have a higher score than a simple answer. Surprisingly, our model can sometimes even generate better answers than those provided, such as `Just the legs of someone' vs. `Part of a person' in the bottom-left image and `behind the truck' vs. `In the background' in the bottom-right image.

\vspace{-5pt}
\section{Conclusions}
The limitations of previous input-answer fusion methods mean that they cannot correctly represent the common vector of these features. As a result, they omit detailed information and focus on short and safe answers. In this paper, we develop a synergistic network that jointly learns the representation of the image, question, answer, and history in a single step. We also improve the N-pair loss function to solve the class-imbalanced problem in the discriminative model. Our final proposed discriminative model achieves state-of-the-art performance on the Visual Dialog v1.0 test-standard server with the robust NDCG evaluation metric. The results of our generative model are also encouraging.

\vspace{-5pt}
\section{Acknowledgement}
This work was supported in part by the Australian Research Council under Project FL-170100117, Project DP-180103424, Project IH180100002, and Project DE-180101438.
{\small
\bibliographystyle{ieee}
\bibliography{egbib}
}

\end{document}